\documentclass[11pt]{article}

\usepackage{acl}
\usepackage{times}
\usepackage{latexsym}
\usepackage[T1]{fontenc}
\usepackage[utf8]{inputenc}
\usepackage{microtype}
\usepackage{inconsolata}
\usepackage{graphicx}
\usepackage{float}
\usepackage{booktabs}
\usepackage{colortbl}
\usepackage[table]{xcolor} 
\usepackage{multirow}     
\usepackage{array}

\definecolor{blue3}{RGB}{180,215,255}  
\definecolor{blue2}{RGB}{120,185,255}  
\definecolor{blue1}{RGB}{70,140,230}    
\definecolor{lingoHL}{RGB}{255,245,180} 

\title{Lingo\_Research\_Group at SemEval-2026 Task 9: Evaluating Prompt Variants for Polarization Detection}

\author{
\textbf{Pritam Kadasi\textsuperscript{1}\thanks{Equal contribution. Corresponding author.}} \quad
\textbf{Anuj Tiwari\textsuperscript{2}\thanks{Equal contribution.}} \quad
Mayank Singh\textsuperscript{1} \\
\textsuperscript{1}Lingo Research Group, Indian Institute of Technology Gandhinagar \\
\textsuperscript{2}Noida Institute of Engineering and Technology 
\textsuperscript{2}ML Collective \\
\small{\texttt{aj11anuj123@gmail.com}}, \\
\small{\texttt{\{pritam.k, singh.mayank\}@iitgn.ac.in}} 
}

\begin{document}
\maketitle

\begin{abstract}
Our submission presented in this paper is for SemEval-2026 Task 9: Multilingual Text Classification Challenge - Polarization Detection and it covers all three subtasks: (1) binary polarization detection, (2) polarization type classification and (3) polarization manifestation identification. We adopt a systematic approach of research on short designed prompts by considering twelve designed prompts that are different in terminology clarity, detail of the definition, guidance of reasoning and in-context examples use. The experiments are conducted using aya-101 and Gemma3-27B, with the latter chosen for the submission at the end of the development through performance considerations. Our system has an average macro level \textbf{F1-score of 0.762 on Subtask 1, 0.587 on Subtask 2 and 0.444 on Subtask 3} with the average accuracy of 0.819, 0.678 and 0.498, respectively, on the official test set averaged among 22 languages, respectively. With cross-task and cross-lingual analysis, we demonstrate that prompt-based approaches can be used effectively to detect coarse-grained polarization but encounter more and more difficulties as far as fine-grained and multi-label sociolinguistic classification is concerned.
\end{abstract}

\section{Introduction}

The issue of online polarization has become a characteristic element of socio-political speech nowadays \cite{10835788}. Social media platforms are becoming more dominated by content presented in antagonistic in-group and out-group terms, and frequently backed by animosity, ostracism, or rhetorical dismissal. The use of such language can enhance social separatism and destroy the constructive social discourse, and automatic detecting and analyzing polarization is therefore a valuable issue of natural language processing (NLP), especially in multilingual and cross-cultural settings. 

Task 9 of SemEval-2026 deals with this issue by introducing the proposal of a multilingual benchmark of polarization analysis, which relates to various languages, events and socio-political contexts. The task is divided into three subtasks that have a higher level of analysis.
\begin{itemize}
    \item Subtask 1 is aimed at the identification of the presence of attitude polarization in a text.
    \item Subtask 2 is devoted to the determination of polarization targets.
    \item Subtask 3 specifies the way polarization is conveyed in certain rhetorical ways.
\end{itemize}

The detection of polarization is a difficult task since this process is mostly realized through framing, rhetoric, or cultural but not hate speech references. Such difficulties are multiplied in multilingual environments, where language structure and discourse patterns differ a lot. Even though task-based off-the-shelf methodologies grounded on task-specific finetuning have been found to be successful \cite{lester-etal-2021-power}, they are commonly limited by the cost of annotation, data space, as well as scalability to low-resource languages.

Most recent developments in large language models (LLMs) have demonstrated that prompt-based inference may be useful in complex classification without finetuning \citep{brown2020}. We analyze prompt-based methods on each of the 3 subsets of SemEval-2026 Task 9 \cite{naseem2026polarbenchmarkmultilingualmulticultural}, analyzing twelve prompt variations without parameter adjustments. The experiments of the aya-101 and Gemma3-27B check the interaction of prompt design and the complexity of the tasks and language variations that provide the advantages and shortcomings of prompt-based multilingual polarization analysis.

\section{Task Description}

\subsection{Subtask 1: Polarization Detection}

In Subtask 1, a binary classification task is performed, where one should answer whether a certain post on social media was attitude polarization. One declares a text polarized, when the text conveys divisive in-group versus out-group framing, hostility, intolerance, or exclusion of another group of people or perspective. It is described through the language and framing of the speaker only and the emotional reaction of the reader is not considered.

\subsection{Subtask 2: Polarization Classification}

Subtask 2 is a continuation of Subtask 1, but it determines the objects of polarization in polarized text. It is a multi label classification assignment, each reading can be put in one or more of the following categories: political, racial/ethnic, religious, gender/sexual, or other. A text can have a combination of more than two polarization targets.

\subsection{Subtask 3: Manifestation Identification}

In subtask 3, the analysis is further narrowed by identifying the expression of polarization in a text. It is also a multi-label classification problem whereby systems need to recognize one or more of the following manifestations: stereotype, vilification, dehumanization, extreme language, lack of empathy and invalidation.

\subsection{Dataset and Evaluation}

The data collection of the posts in a social media that are based on various sociopolitical settings and events, such as elections, clashes, demonstrations, and social conversations were used which was provided by SemEval. The assignment discusses several languages such as Amharic, Arabic, Bengali, Burmese, Chinese, English, German, Hausa, Hindi, Italian, Khmer, Nepali, Odia, Persian, Punjabi, Russian, Spanish, Swahili, Telugu, Turkish, and Urdu. Not all languages have all their subtasks or labels because of some variations in the scope of annotation and access to data. To measure system performance in all subtasks, the use of macro-averaged F1-score is used where all labels are considered and is appropriate because of the presence of a class imbalance. It is presented with labeled training data, and it is lastly evaluated on a held-out test set.

\section{Methodology}

All three subtasks of SemEval-2026 task 9 are solved with a prompt based inference \cite{schick-schutze-2021-exploiting} system on multilingual LLMs. We formulate polarization analysis as an instruction following problem and study how systematic variations in prompt design affect performance across tasks and languages. The hypothesis that polarization and particularly its targets and manifestation is a context dependent and implicitly realized phenomenon, and that the performance of models is heavily dependent upon the articulateness of the specification of task definitions, decision limits, and reasoning expectations in the prompt guides our methodology.

\subsection{Problem Formulation}
Social media post x: the purpose varies by subtask:

\begin{itemize}
    \item Subtask 1. Predict a binary label y, which is in the range of 0,1, on whether the text is attitude polarizing or not.
    \item Subtask 2: Predict a multi-label vector indicating the presence or absence of polarization toward each target category.
    \item Subtask 3: Make a guess of a multi-labeled vector of the rhetorical forms of polarization.
\end{itemize}

\subsection{Prompt Design as Control Experimental variable}

Rather than using a single prompt, we come up with twelve variants of prompts that build up to more specify the task, more detail of its context, and reasoning instructions. We started with most basic fundamental prompt and kept adding one variable to it to create a new prompt that could likely affect performance, then via trial and testing we crafted these 12 prompts. The questions are grouped together in a conceptual gradient, and we can examine the effects of the various kinds of instruction on the model behavior. The entire prompt set is given in the appendix. The major design dimensions have been summarized below.

\begin{itemize}
    \item Prompts 1-2 give minimal or no context of the task.
    \item Prompts 3-4 provide to the short task definition.
    \item Prompts 5-6 have a clear description of polarized and non-polarized cases.
    \item Prompts 7-8 require the model to follow the text step by step before coming up with a final prediction.
    \item Prompts 9-12 bring about in-context sample.
\end{itemize}

We evaluate all prompt variants using two multilingual LLMs: aya-101 \cite{ustun-etal-2024-aya} and Gemma3-27B \cite{gemma_2025} due to their known multilingual abilities in diverse set of languages. Across subtasks and prompt configurations, Gemma3-27B outperforms macro F1-scores across subtasks and prompts with varied degrees of richness, So, Gemma3-27B is chosen.

\subsection{Inference Procedure}

In the case of every input instance, the chosen prompt is appended with the original text and fed into the model in an inference. In Subtask 1, the model results in a single binary classification. On the case of Subtasks 2 and 3, the model will generate a list of binary choices that will be associated with each target or manifestation label. The most instructive prompt with clear tasks and illustrative examples is chosen to remain the same in all the submissions of the test.

\section{Experimental Setup}

Our experiments are performed on datasets published on SemEval-2026 Task 9 \cite{naseem-etal-2026-polar}. All subtasks have labeled training data, which is applied to do prompt evaluation and selection. None of the external data and augmentation is used. The organizers evaluate held-out test sets and do official evaluation.

Macro-averaged F1-score is used to measure system performance as task organizers suggest. Macro F1 weighs all the labels which are equal and thus, it is suitable in addressing the issue of class imbalance in both binary (Subtask 1) and multi-label (Subtasks 2 and 3) tasks. Selection is done of prompts on the basis of changes in training-time performance, and test labels are not available. Prompts are in English and have been applied the same way across languages, depending on the multilingual abilities of the models. Language-specific custom prompting is not applied.

All experiments used fixed generation configuration to ensure consistency across prompt variants. In the case of Subtask 1, the output of the model is one binary label. In case of Subtasks 2 and 3, the model provides a collection of binary predictions to each of the target or manifestation labels.

\section{Results}

This section gives the official results of the evaluation of our system on SemEval-2026 Task 9, including Subtasks 1-3.

\begin{figure}[htbp]
    \centering
    \includegraphics[width=1\linewidth]{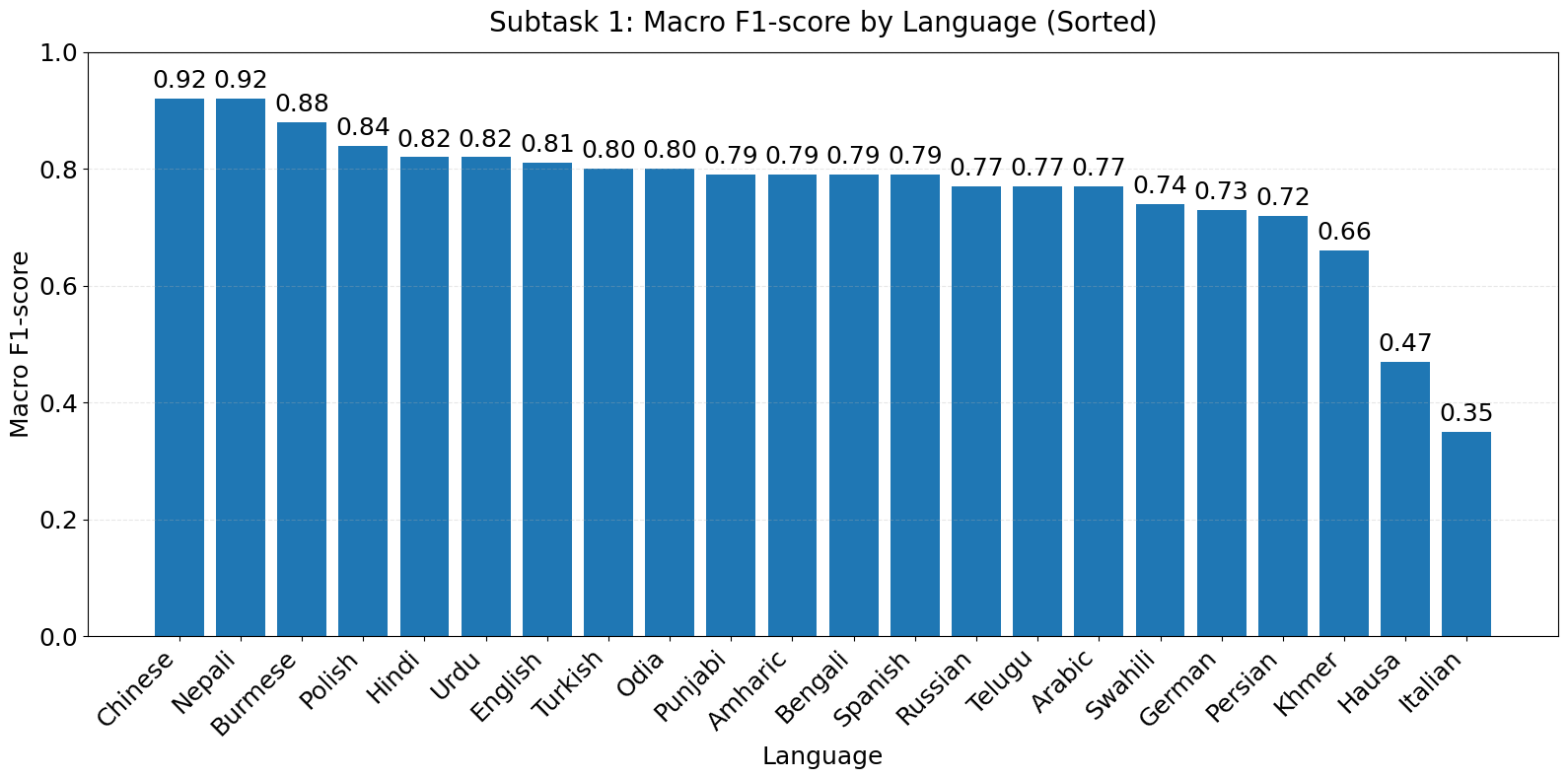}
    \caption{Macro averaged F1-scores for Subtask 1 across 22 languages sorted performance wise. Performance varies across languages from 0.92 for Chinese and Nepali to 0.35 for the Italian language.}
    \label{fig:placeholder}
\end{figure}

Figure 1 gives a summation of macro F1-scores of 22 languages under Subtask 1. The best system, which is based on Gemma3-27B and the chosen prompt, has an average \textbf{macro F1-score of 0.762} and an \textbf{average accuracy of 0.819} and, therefore, is very promising in terms of binary polarization detection in a multilingual environment. The complete per-language macro F1-scores and accuracies are reported in Table 1 and Table 2 (Appendix).

In Subtask 2, the system has an \textbf{average macro F1-score of 0.587} on 22 languages and Subtask 3 has an \textbf{average macro F1-score of 0.444} on 18 languages. This gradual reduction is indicative of the growing complexity of the tasks which no longer are binary detected but multi-label discriminated with respect to abstract sociolinguistic types.

\begin{figure}[htbp]
    \centering
    \includegraphics[width=1\linewidth]{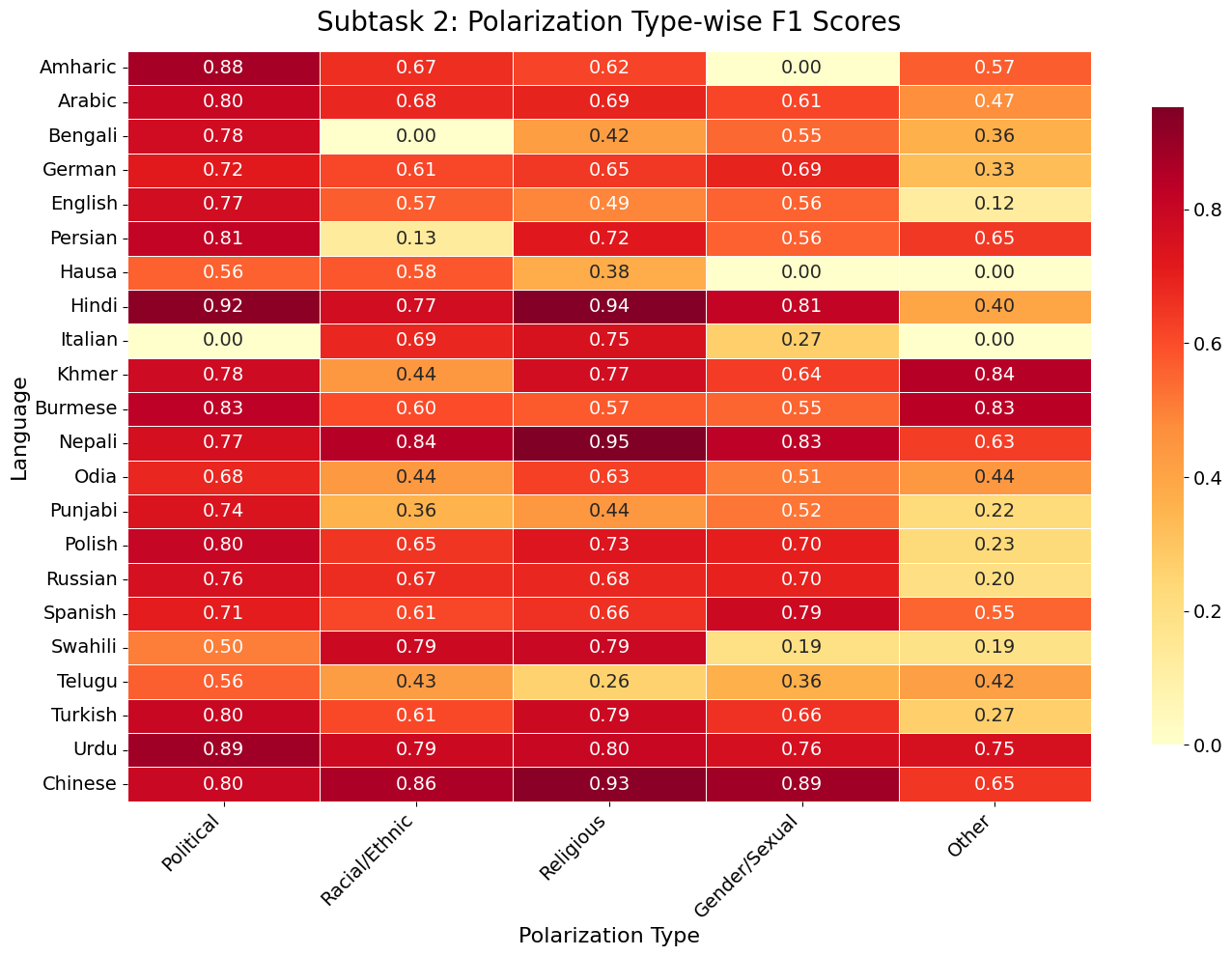}
    \caption{Macro-averaged F1-scores of Subtask 2 by polarization categories and in 22 languages.}
    \label{fig:placeholder}
\end{figure}

\begin{figure}[htbp]
    \centering
    \includegraphics[width=1\linewidth]{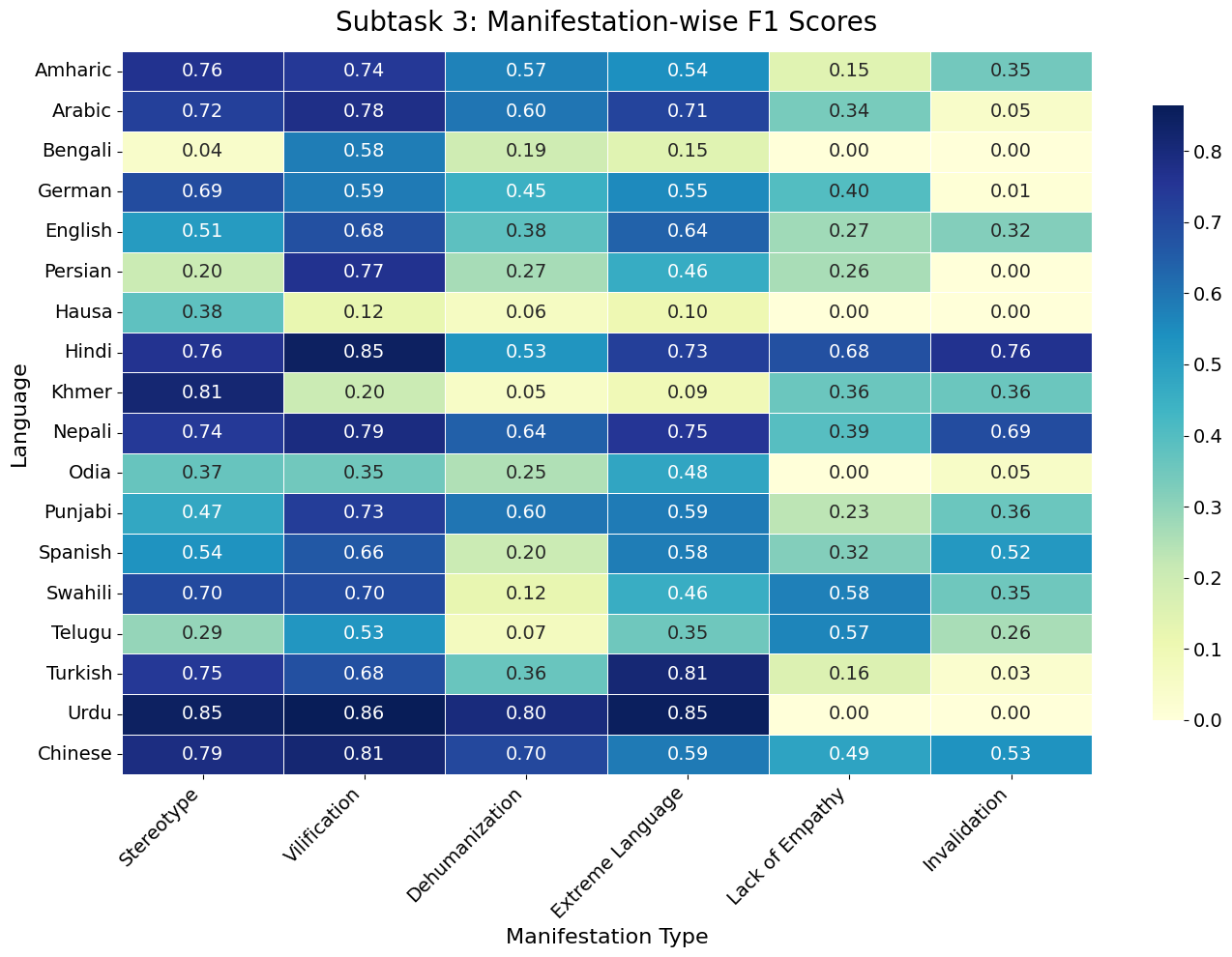}
    \caption{F1-scores of Subtask 3 by manifestation categories of F1 and 18 languages, macro- averaged.}
    \label{fig:placeholder}
\end{figure}

Figure 2 is a heatmap of target wise F1-scores of Subtask 2, showing a significant range of values between languages and polarization targets. The per-category breakdown across all 22 languages is provided in Table 3 (Appendix). Figure 3 represents the heatmap of Subtask 3, which has the lowest average overall performance of the three subtasks, which normally have low F1-scores in multiple languages. Full per-language metrics for Subtask 3 are reported in Table 4 (Appendix).

Overall, the findings demonstrate that prompt-based multilingual LLMs are useful in coarse grained polarization detection but that responding more finely, in tasks that involve fine-grained multi-label sociolinguistic inferences, presents growing challenges. Figures 1-3 reveal that task complexity, label abstraction and cross-lingual variation together affect performance. The results with fully per language and per label items on all subtasks are entirely presented in the Appendix.

\section{Analysis}

\subsection{Cross-Subtask Degradation of performance}

The major trend in the results is a systematic decrease in performance between Subtask 1 and Subtask 3. \textbf{Although Subtask 1 has a mean macro F1-score of 0.762} in 22 languages, this reduces to 0.587 in Subtask 2 and further to 0.444 in Subtask 3 as does the accuracy. This degradation is explained as a result of three factors playing out.

\begin{itemize}
    \item Subtask 1 involves one decision regarding the presence of polarization but Subtask 2 and 3 involve prediction of several labels that might overlap. Macro F1 is the error rate of a single label, therefore the rate of errors grows exponentially with the dimensionality of labels.
    \item Subtask 2 involves definition of objects of polarization (e.g., political, religious), as the latter are likely to be lexically indicated. However, in Subtask 3, the manifestations that need to be identified include lack of empathy or invalidation, which might hardly be denoted by clear indicators therefore significantly more difficult to be zero-shot prompted.
    \item As indicated in the appendix tables, the frequency of several labels of Subtask 3 vary among languages. Removal of a single positive case causes drastic decreases in macro F1, which explains why it is only one language that has a \textbf{macro F1 above 0.70} in Subtask 3.
\end{itemize}

\subsection{Error Patterns and Recall Suppression Across Languages}

We find the heterogeneous performance across languages with each subtask and, to a large extent is due to differences in recall as opposed to precision. Some of the languages show high accuracy and low macro F1, respectively, which reflects the systematic under-prediction of positive labels. Hausa, Italian, and Khmer, are some of the languages that depict this trend. Polarized instances are often overlooked in such cases and especially when they are indirectly conveyed by sarcasm or rhetorical inquiries or by reference to culture-specific situations. The behaviour indicates that in absence of specific hostility or group enmity, the model will be reduced to the non-polarized label.

This inhibition of recall is enhanced in Subtasks 2 and 3 where the model is required to recognize particular targets or manifestations as opposed to polarization in the broadest sense. Consequently, the accuracy will be misleadingly high but macro F1 will decrease dramatically.

\subsection{Effect of Conservative Prompting}

The optimum prompt among subtasks is a conservative decision maker, which is to only predict polarization when there is explicit and unambiguous evidence. This works well in the case of Subtask 1 where it minimizes the false positives and gives high macro F1.

Nevertheless, in the Subtasks 2 and 3, the identical strategy is becoming more expensive. Conservative prompting induces blocking of subtle targets and manifestations especially in languages where polarization is implicitly manifested or which depend on common cultural ground. This trade-off is the reason why the accuracy is relatively high despite the fact that macro F1 drops significantly between the Subtask 1 and Subtask 3.

\subsection{Language-Specific Strengths Across Subtasks}

The other interesting point that could be noted is that languages predominate over various subtasks. In Subtasks 1 and 2, Nepali and Chinese have relatively good overall F1 scores, whereas in Subtask 3, Hindi has the best F1 score. This trend can be attributed to dissimilarity in discourse feature and compatibility with task-specifications. Polarization in Nepali and Chinese is commonly formulated by the explicit use of group reference and topical markers which suffice to deal in binary recognition and targeting. Consequently, both languages do well in Subtasks 1 and 2. 

Contrary to Hindi, Hindi involves more apparent rhetorical tactics, by vilifying, using extreme language, and directly invalidating which directly translate into labels in Subtask 3 manifestations. Moreover, the Hindi sociopolitical discourse has good handle in multilingual model pretraining offering more power priors to manifestation level of reasoning. These findings indicate that polarization is conventionally reflected in terms of language resources instead of the general availability of the resources.

\subsection{Why English Underperforms}

English tends to show a performance at or below the median in the subtasks although much pretraining has been made on the English data. 

English polarization is often based on irony, sarcasm, rhetorical framing and ideological shorthand. The terms may not be directly hostile or even referring to groups, and as a result, conservative prompts would describe them with the designation of not being polarized. For instance, consider statements like \textbf{"A small price to pay for Ukrainian sovereignty and our green future. Stop whining,"} which uses sarcasm to dismiss legitimate concerns without explicit hostility; \textbf{"They are now the deep state,"} which relies on ideological shorthand rather than overt group-based attacks; or \textbf{"All in the name of Jesus, the ultimate socialist,"} where political and religious targets are fused through ironic reframing. In each case, our model misclassified these as non-polarized because the surface-level text lacks the direct hostility, us-versus-them language, or dehumanization. However, non-English datasets tend to have a higher amount of direct and explicit polarization cues, which can be discovered by prompt-based frameworks more readily.

\subsection{Impact of Translation-Based Reasoning}

Prompts asking the model to encode inputs in English and only then the inputs are classified produce more consistent behavior on certain languages, which probably are better explained by more internalized reasoning in English. The translation may however also defuse emotionally charged expressions. This underscores one of the fundamental shortcomings of translation-oriented prompting, i.e. semantic equivalence does not imply maintaining the semantically parallel rhetorical intent, which is at the core of manifestation identification.

Due to the predominance of languages with English or Western contexts, examples may miss forms of polarization specific to a region of the world, which constrains generalization. This indicates that the example diversity is just as significant as the example presence in multilingual sociopolitical activities.

\section{Limitations}

Our preference of conservative decision-making strategies is systematic in that it gives precision a stronger edge than recall. We also didn't tried applying these prompts in other languages natively other than in English, future work could explore that aspect as well.

\section{Conclusion}

In this paper, a prompt-based model in the SemEval-2026 Task 9 the detection of polarization, the classification of types and the recognition of manifestation have been described. In the absence of task-based finetuning, we presented all of our subtasks as problems of instruction following, and methodically tested prompt design with multilingual LLMs. In general, the findings outline the advances as well as shortcomings of the prompt-based method to the multilingual sociopolitical text analysis, where performance depends on the task difficulty, labeling abstraction, and discourse versatility.

\section{Ethical Considerations}

\begin{itemize}
    \item \textbf{Data Source and Privacy}: The dataset is being provided in SemEval-2026 Task 9 and is represented by publicly available social media posts labeled with polarization, its targets, and manifestations. We did not gather new user information or even make any effort to recognize, trace, or profile persons.
    \item \textbf{Risk of Misclassification}: False positive can be a factor in unfair moderation or even banning of legal speech. False negatives can be used to tolerate dangerous or divisive rhetoric. We explicitly caution that such systems should not be used as standalone moderation tools without human oversight.
    \item \textbf{Bias and Cross-Lingual Disparities}: The obtained outcomes indicate that there is significant performance difference among the languages. Such differences can be due to differences in the representations of pretraining data, Cultural and rhetorical variation, Annotation density and class imbalance, Translation-based reasoning effects.
    \item \textbf{Example-Based Representation Bias and Prompting}: Example-augmented prompts (Prompts 9-12) improve performance but may introduce representational bias.
\end{itemize}

\bibliography{custom}

\appendix
\newpage

\section{Appendix - Full Results}

\begin{table}[H]
\centering
\small
\caption{The result of Subtask 1 binary polarization detection, as a measure of performance in Subtask 1 of 22 languages}
\resizebox{\columnwidth}{!}{%
\begin{tabular}{lcccccc}
\toprule
\textbf{Language} & \textbf{Accuracy} & \textbf{Precision} & \textbf{Recall} & \textbf{F1 Binary} & \textbf{F1 Macro} & \textbf{F1 Micro} \\
\midrule
Amharic   & 0.8488 & 0.8748 & 0.9277 & 0.9005 & 0.7928 & 0.8488 \\
Arabic    & 0.7857 & 0.9118 & 0.5771 & 0.7068 & 0.7690 & 0.7857 \\
Bengali   & 0.8048 & 0.8386 & 0.6651 & 0.7419 & 0.7925 & 0.8048 \\
German    & 0.7367 & 0.7764 & 0.6327 & 0.6972 & 0.7322 & 0.7367 \\
English   & 0.8264 & 0.7873 & 0.7223 & 0.7534 & 0.8098 & 0.8264 \\
Persian   & 0.7668 & 0.8776 & 0.7962 & 0.8349 & 0.7191 & 0.7668 \\
Hausa     & 0.8936 & 0.0000 & 0.0000 & 0.0000 & 0.4719 & 0.8936 \\
Hindi     & 0.9110 & 0.9443 & 0.9515 & 0.9479 & 0.8212 & 0.9110 \\
Italian   & 0.5260 & 0.3333 & 0.0014 & 0.0027 & 0.3459 & 0.5260 \\
Khmer     & 0.9123 & 0.9307 & 0.9760 & 0.9528 & 0.6645 & 0.9123 \\
Burmese   & 0.8801 & 0.8911 & 0.9007 & 0.8959 & 0.8773 & 0.8801 \\
Nepali    & 0.9181 & 0.9294 & 0.9047 & 0.9169 & 0.9180 & 0.9181 \\
Odia      & 0.8424 & 0.7586 & 0.6535 & 0.7021 & 0.7975 & 0.8424 \\
Punjabi   & 0.7948 & 0.7918 & 0.7837 & 0.7877 & 0.7946 & 0.7948 \\
Polish    & 0.8496 & 0.8550 & 0.7716 & 0.8112 & 0.8431 & 0.8496 \\
Russian   & 0.8329 & 0.8536 & 0.5311 & 0.6548 & 0.7723 & 0.8329 \\
Spanish   & 0.7883 & 0.7830 & 0.7905 & 0.7867 & 0.7883 & 0.7883 \\
Swahili   & 0.7515 & 0.9070 & 0.5624 & 0.6943 & 0.7425 & 0.7515 \\
Telugu    & 0.7786 & 0.9907 & 0.5779 & 0.7300 & 0.7712 & 0.7786 \\
Turkish   & 0.7978 & 0.8480 & 0.7452 & 0.7933 & 0.7977 & 0.7978 \\
Urdu      & 0.8412 & 0.8937 & 0.8752 & 0.8844 & 0.8156 & 0.8412 \\
Chinese   & 0.9211 & 0.9163 & 0.9294 & 0.9228 & 0.9211 & 0.9211 \\
\bottomrule
\end{tabular}
}
\end{table}

\begin{table}[H]
\centering
\small
\caption{It is a table for Multi-label polarization detection and its findings indicate that there is a significant cross-lingual difference, similar to the way that Chinese, Urdu, and Hindi languages perform quite well in macro F1-scores}
\resizebox{\columnwidth}{!}{%
\begin{tabular}{lcccccc}
\toprule
\textbf{Language} & \textbf{F1 Micro} & \textbf{F1 Macro} & \textbf{Precision Micro} & \textbf{Precision Macro} & \textbf{Recall Micro} & \textbf{Recall Macro} \\
\midrule
Amharic   & 0.7680 & 0.5458 & 0.7412 & 0.5507 & 0.7967 & 0.5493 \\
Arabic    & 0.6750 & 0.6517 & 0.6513 & 0.6374 & 0.7005 & 0.6858 \\
Bengali   & 0.6836 & 0.4216 & 0.6841 & 0.5369 & 0.6831 & 0.3766 \\
German    & 0.6373 & 0.5994 & 0.5661 & 0.5535 & 0.7288 & 0.6722 \\
English   & 0.6808 & 0.5027 & 0.6325 & 0.4971 & 0.7370 & 0.5322 \\
Persian   & 0.7266 & 0.5757 & 0.6842 & 0.5588 & 0.7746 & 0.5988 \\
Hausa     & 0.4924 & 0.3022 & 0.5956 & 0.3574 & 0.4197 & 0.2660 \\
Hindi     & 0.8800 & 0.7704 & 0.8573 & 0.7683 & 0.9039 & 0.7850 \\
Italian   & 0.2737 & 0.3409 & 0.6610 & 0.3802 & 0.1726 & 0.3120 \\
Khmer     & 0.8211 & 0.6939 & 0.7372 & 0.6984 & 0.9266 & 0.7058 \\
Burmese   & 0.7755 & 0.6772 & 0.7623 & 0.6668 & 0.7892 & 0.6900 \\
Nepali    & 0.7792 & 0.8047 & 0.7054 & 0.7534 & 0.8703 & 0.8823 \\
Odia      & 0.6119 & 0.5394 & 0.6119 & 0.5654 & 0.6119 & 0.5221 \\
Punjabi   & 0.5395 & 0.4555 & 0.4617 & 0.3855 & 0.6488 & 0.5859 \\
Polish    & 0.7286 & 0.6253 & 0.7525 & 0.7053 & 0.7062 & 0.5996 \\
Russian   & 0.6895 & 0.6023 & 0.6737 & 0.6115 & 0.7061 & 0.6191 \\
Spanish   & 0.6699 & 0.6640 & 0.6172 & 0.6178 & 0.7323 & 0.7226 \\
Swahili   & 0.6785 & 0.4929 & 0.8059 & 0.6157 & 0.5859 & 0.4240 \\
Telugu    & 0.4479 & 0.4054 & 0.4292 & 0.4301 & 0.4683 & 0.4210 \\
Turkish   & 0.7195 & 0.6242 & 0.6768 & 0.6300 & 0.7680 & 0.6571 \\
Urdu      & 0.8002 & 0.7978 & 0.6998 & 0.6994 & 0.9342 & 0.9339 \\
Chinese   & 0.8311 & 0.8250 & 0.7715 & 0.7868 & 0.9007 & 0.8763 \\
\bottomrule
\end{tabular}
}
\end{table}

\begin{table}[H]
\centering
\small
\caption{Macro F1-scores of Subtask 2 by category wise. The performances, both Political and Religious types, tend to be stronger in most languages}
\resizebox{\columnwidth}{!}{%
\begin{tabular}{lccccc}
\toprule
\textbf{Language} & \textbf{Political} & \textbf{Racial/Ethnic} & \textbf{Religious} & \textbf{Gender/Sexual} & \textbf{Other} \\
\midrule
Amharic  & 0.8763 & 0.6675 & 0.6182 & 0.0000 & 0.5672 \\
Arabic   & 0.7970 & 0.6813 & 0.6940 & 0.6149 & 0.4714 \\
Bengali  & 0.7757 & 0.0000 & 0.4231 & 0.5455 & 0.3636 \\
German   & 0.7178 & 0.6136 & 0.6455 & 0.6936 & 0.3266 \\
English  & 0.7736 & 0.5667 & 0.4909 & 0.5574 & 0.1250 \\
Persian  & 0.8139 & 0.1333 & 0.7248 & 0.5600 & 0.6463 \\
Hausa    & 0.5584 & 0.5778 & 0.3750 & 0.0000 & 0.0000 \\
Hindi    & 0.9243 & 0.7734 & 0.9426 & 0.8116 & 0.4000 \\
Italian  & 0.0000 & 0.6855 & 0.7500 & 0.2692 & 0.0000 \\
Khmer    & 0.7814 & 0.4416 & 0.7670 & 0.6364 & 0.8432 \\
Burmese  & 0.8265 & 0.6000 & 0.5747 & 0.5512 & 0.8336 \\
Nepali   & 0.7658 & 0.8436 & 0.9524 & 0.8283 & 0.6333 \\
Odia     & 0.6823 & 0.4375 & 0.6286 & 0.5075 & 0.4412 \\
Punjabi  & 0.7417 & 0.3553 & 0.4410 & 0.5198 & 0.2195 \\
Polish   & 0.8040 & 0.6537 & 0.7342 & 0.7045 & 0.2299 \\
Russian  & 0.7564 & 0.6730 & 0.6822 & 0.6957 & 0.2041 \\
Spanish  & 0.7088 & 0.6119 & 0.6613 & 0.7867 & 0.5515 \\
Swahili  & 0.5038 & 0.7852 & 0.7943 & 0.1942 & 0.1871 \\
Telugu   & 0.5621 & 0.4259 & 0.2571 & 0.3644 & 0.4173 \\
Turkish  & 0.7968 & 0.6083 & 0.7866 & 0.6598 & 0.2692 \\
Urdu     & 0.8874 & 0.7861 & 0.7983 & 0.7624 & 0.7549 \\
Chinese  & 0.7953 & 0.8639 & 0.9296 & 0.8857 & 0.6508 \\
\bottomrule
\end{tabular}
}
\end{table}

\begin{table}[H]
\centering
\small
\caption{Subtask 3 performance rates of 18 languages}
\resizebox{\columnwidth}{!}{%
\begin{tabular}{lccccccc}
\toprule
\textbf{Language} & \textbf{F1 Micro} & \textbf{F1 Macro} & \textbf{Precision Micro} & \textbf{Precision Macro} & \textbf{Recall Micro} & \textbf{Recall Macro} & \textbf{Exact Match} \\
\midrule
Amharic  & 0.6417 & 0.5190 & 0.5588 & 0.5398 & 0.7533 & 0.5968 & 0.2438 \\
Arabic   & 0.6767 & 0.5329 & 0.6071 & 0.5856 & 0.7642 & 0.5997 & 0.4839 \\
Bengali  & 0.4353 & 0.1609 & 0.4503 & 0.2789 & 0.4213 & 0.1713 & 0.6556 \\
German   & 0.5416 & 0.4494 & 0.4666 & 0.4280 & 0.6455 & 0.5761 & 0.3652 \\
English  & 0.5296 & 0.4686 & 0.4887 & 0.4409 & 0.5779 & 0.5170 & 0.5696 \\
Persian  & 0.5875 & 0.3259 & 0.5836 & 0.4066 & 0.5915 & 0.3303 & 0.4259 \\
Hausa    & 0.2281 & 0.1111 & 0.3171 & 0.2211 & 0.1781 & 0.1076 & 0.8905 \\
Hindi    & 0.7504 & 0.7186 & 0.6994 & 0.7015 & 0.8094 & 0.7631 & 0.2403 \\
Khmer    & 0.7156 & 0.3136 & 0.6339 & 0.4773 & 0.8215 & 0.3472 & 0.6426 \\
Nepali   & 0.7230 & 0.6685 & 0.6672 & 0.6538 & 0.7891 & 0.7078 & 0.5836 \\
Odia     & 0.3810 & 0.2490 & 0.3777 & 0.3850 & 0.3842 & 0.2445 & 0.6857 \\
Punjabi  & 0.5653 & 0.4972 & 0.5200 & 0.4810 & 0.6193 & 0.5656 & 0.4722 \\
Spanish  & 0.5270 & 0.4706 & 0.4734 & 0.4648 & 0.5943 & 0.5197 & 0.4126 \\
Swahili  & 0.5679 & 0.4844 & 0.5433 & 0.4750 & 0.5949 & 0.5147 & 0.4210 \\
Telugu   & 0.4357 & 0.3451 & 0.4156 & 0.4134 & 0.4579 & 0.3796 & 0.4765 \\
Turkish  & 0.6924 & 0.4656 & 0.6378 & 0.5490 & 0.7572 & 0.5130 & 0.4968 \\
Urdu     & 0.6949 & 0.5606 & 0.7678 & 0.8452 & 0.6345 & 0.6194 & 0.2111 \\
Chinese  & 0.7303 & 0.6535 & 0.6918 & 0.6550 & 0.7734 & 0.6692 & 0.6902 \\
\bottomrule
\end{tabular}
}
\end{table}

\begin{table}[H]
\centering
\small
\caption{Category wise macro F1-scores of Subtask 3, in six manifestation types (Stereotype, Vilification, Dehumanization, Extreme Language, Lack of Empathy and invalidation)}
\resizebox{\columnwidth}{!}{%
\begin{tabular}{lcccccc}
\toprule
\textbf{Language} & \textbf{Stereotype} & \textbf{Vilification} & \textbf{Dehumanization} & \textbf{Extreme Language} & \textbf{Lack of Empathy} & \textbf{Invalidation} \\
\midrule
Amharic  & 0.7632 & 0.7449 & 0.5714 & 0.5428 & 0.1463 & 0.3452 \\
Arabic   & 0.7225 & 0.7780 & 0.6000 & 0.7142 & 0.3365 & 0.0465 \\
Bengali  & 0.0417 & 0.5828 & 0.1930 & 0.1481 & 0.0000 & 0.0000 \\
German   & 0.6948 & 0.5900 & 0.4498 & 0.5542 & 0.3990 & 0.0085 \\
English  & 0.5109 & 0.6844 & 0.3815 & 0.6402 & 0.2745 & 0.3198 \\
Persian  & 0.2027 & 0.7653 & 0.2655 & 0.4603 & 0.2614 & 0.0000 \\
Hausa    & 0.3810 & 0.1212 & 0.0625 & 0.1017 & 0.0000 & 0.0000 \\
Hindi    & 0.7613 & 0.8469 & 0.5269 & 0.7275 & 0.6844 & 0.7646 \\
Khmer    & 0.8139 & 0.2029 & 0.0526 & 0.0920 & 0.3605 & 0.3596 \\
Nepali   & 0.7394 & 0.7918 & 0.6435 & 0.7505 & 0.3924 & 0.6935 \\
Odia     & 0.3677 & 0.3478 & 0.2500 & 0.4798 & 0.0000 & 0.0488 \\
Punjabi  & 0.4728 & 0.7323 & 0.6004 & 0.5871 & 0.2308 & 0.3600 \\
Spanish  & 0.5358 & 0.6633 & 0.2043 & 0.5828 & 0.3193 & 0.5182 \\
Swahili  & 0.6993 & 0.6955 & 0.1244 & 0.4571 & 0.5756 & 0.3543 \\
Telugu   & 0.2920 & 0.5253 & 0.0714 & 0.3518 & 0.5667 & 0.2632 \\
Turkish  & 0.7454 & 0.6844 & 0.3636 & 0.8108 & 0.1566 & 0.0328 \\
Urdu     & 0.8465 & 0.8642 & 0.7969 & 0.8518 & 0.0022 & 0.0022 \\
Chinese  & 0.7894 & 0.8145 & 0.7048 & 0.5891 & 0.4884 & 0.5348 \\
\bottomrule
\end{tabular}
}
\end{table}

\section{Appendix - Prompt Design Details}
This appendix provides the full set of prompt variants used in our experiments, along with a brief description of their design intent.

\subsection{Minimal Instruction Prompts}
Prompt 1: This prompt provides no task description, no definition of polarization, and no language information testing whether polarization can be inferred without explicit guidance.
{\small
\begin{verbatim}
PROMPT = """
Classify the following text as 0 or 1: {text} 
Output only 0 or 1:
"""
\end{verbatim}
}

Prompt 2: This prompt introduces explicit language identification (e.g., specifying that the input text is in Amharic) and clarifies that the task is binary classification.
{\small
\begin{verbatim}
PROMPT = """
The text is in Amharic, You are a text 
classification model. Classify the following 
Amharic text as either 0 or 1. 
Text: {text} 
Output only the number (0 or 1):
"""
\end{verbatim}
}

\subsection{Basic Task Definition Prompts}
Prompt 3: This prompt adds a concise task definition but the instruction remains brief with minimal contextual elaboration.
{\small
\begin{verbatim}
PROMPT = """
Is the following text polarized? 
(Polarized = hostile or divisive) 
Return only "1" if polarized else "0". 
Text: {TEXT}
"""
\end{verbatim}
}

Prompt 4: This prompt instructs to first translate the input text into English and then perform polarization classification, leveraging the model’s stronger reasoning capabilities in English.
{\small
\begin{verbatim}
PROMPT = """
Task: Translate and Classify for 
Polarization Language:
Step 1: Translate the input text into 
fluent, natural English.
Step 2: Decide whether the translated 
text expresses polarization: 
(Polarization = hostility, insults, 
dehumanization, or us-vs-them framing).
Step 3: Return a JSON object with the 
English translation and the label.

Output only "1" if polarized else "0".
Text: {TEXT}
"""
\end{verbatim}
}

\subsection{Decision Boundary Clarification Prompts}
Prompt 5: This prompt provides a more detailed description of polarized content.
{\small
\begin{verbatim}
PROMPT = """
You are a content moderator. Task: Does 
this post contain polarized speech?
(Polarized = hostile or divisive language 
directed at groups or protected attributes, 
or calls for exclusion.) Consider:calm critique, 
neutral reporting → 0
- insults, slurs, demeaning metaphors, or 
dehumanizing language → 1
- Return ONLY "1" if polarized else "0".

Post: {TEXT}
"""
\end{verbatim}
}

Prompt 6: This prompt instructs to output the polarized label only when polarization is explicit and unambiguous, and default to non-polarized when uncertain. This prompt is designed to reduce false positives by sharpening the decision boundary.
{\small
\begin{verbatim}
PROMPT = """
Task: Conservative Polarization Detection
Guidance:
- Output 1 ONLY if polarization is explicit, 
clear, and unambiguous.
- If the post expresses disagreement calmly, 
or uses mild criticism without hostility or 
group demeaning, choose 0.
- When in doubt, choose 0. Return exactly "1" 
or "0" with no additional text.

Text: {TEXT}
"""
\end{verbatim}
}

\subsection{Reasoning-guided Prompts}
Prompt 7: This prompt explicitly instructs the model to reason internally about the presence of polarization before producing a final label.
{\small
\begin{verbatim}
PROMPT = """
Task: Polarization Detection (reason then answer) 
Instructions:
1. In your head, consider whether the text shows 
hostility, dehumanization, insults, or explicit 
us-versus-them frames.
2. Then output ONLY a JSON object with the final 
label. Final output only "1" if polarized else "0".
    
Text: {TEXT}
"""
\end{verbatim}
}

Prompt 8: This prompt provides a more descriptive definition of both polarized and non-polarized content, along with step-by-step instructions emphasizing full-context consideration rather than reliance on individual keywords.
{\small
\begin{verbatim}
PROMPT = """
Task: Strict Polarization Detection

Definition: A text is POLARIZED when it clearly 
expresses hostility, antagonism, insults, 
demeaning language, dehumanization, calls for 
exclusion, or explicit us-versus-them framing 
directed at a group or individual. Neutral reporting, 
calm opinion, or factual statements are NOT polarized.

Instructions:
1. Consider the full text and context; do not classify 
based only on single words.
2. Output exactly one number.
3. Use '1' for polarized and '0' for not polarized.
   
Text: {TEXT}
"""
\end{verbatim}
}

\subsection{Example-augmented Prompts}
Prompt 9: This prompt introduces five labeled examples illustrating both polarized and non-polarized cases, following a precise task definition.
{\small
\begin{verbatim}
PROMPT = """
Task: Polarization Detection

Definition: Label a text 1 if it contains explicit 
hostility, insults, dehumanization, or us-vs-them 
framing. Label 0 otherwise.

Examples -
Example 1:
Text: "Do the Democrats actually do anything"
Label: "polarized": 1
   
Example 2:
Text: "Do you think the United States should have 
open borders?"
Label: "non polarized": 0
   
Example 3:
Text: "Does anyone know what time the Ukraine 
invasion World War 3 is supposed to start?"
Label: "non polarized": 0

Example 4:
Text: "Donald Trump is a hoe"
Label: "polarized": 1

Example 5:
Text: "Fake news, begone English propaganda!"
Label: "polarized": 1

Now classify the following. Output ONLY "1" 
if polarized and "0" only if it is non-polarized.
   
Text: {TEXT}
"""
\end{verbatim}
}

Prompt 10: This prompt provides an in-depth conceptual definition of polarization, covering a wide range of aspects such as stereotyping, vilification, and intolerance, but does not include any examples.
{\small
\begin{verbatim}
PROMPT = """
Polarization is the sharp division of opinions 
into opposing groups, often with hostility and 
exclusion. Polarization refers to the process 
or phenomenon in which opinions, beliefs, or 
behaviors become more extreme or divided, 
leading to a greater distance or conflict 
between differing groups. Attitude polarization 
is the negative attitude that individuals or 
groups display towards individuals and groups 
outside their group while also showing blind 
support and solidarity towards people within their 
group. Polarization denotes stereotyping, 
vilification, dehumanization, deindividuation, or 
intolerance of other people's views, beliefs, and 
identities.
   
In this task, speeches and articles that are shared 
on social media that incite division, groupism, 
hatred, conflict, and intolerance are shared. Your 
task is to classify the given text as either '0' for 
non polarized classification or '1' for polarized 
classification.
   
Text: {text}
"""
\end{verbatim}
}

Prompt 11: This prompt combines the detailed conceptual definition with five carefully curated labeled examples. The examples span multiple forms of polarization, including political and ideological cases.
{\small
\begin{verbatim}
PROMPT = """
Polarization is the sharp division of opinions 
into opposing groups, often with hostility and 
exclusion. Polarization refers to the process 
or phenomenon in which opinions, beliefs, or 
behaviors become more extreme or divided, 
leading to a greater distance or conflict 
between differing groups. Attitude polarization 
is the negative attitude that individuals or 
groups display towards individuals and groups 
outside their group while also showing blind 
support and solidarity towards people within 
their group. Polarization denotes stereotyping, 
vilification, dehumanization, deindividuation, 
or intolerance of other people's views, beliefs, 
and identities.
   
In this task, speeches and articles that are 
shared on social media that incite division, 
groupism, hatred, conflict, and intolerance are 
shared. Your task is to classify the given text 
as either '0' for non polarized classification 
or '1' for polarized classification.

Examples -
Example 1:
Text: "Do the Democrats actually do anything"
Label: "polarized": 1
   
Example 2:
Text: "Do you think the United States should have 
open borders?"
Label: "non polarized": 0
   
Example 3:
Text: "Does anyone know what time the Ukraine 
invasion World War 3 is supposed to start?"
Label: "non polarized": 0

Example 4:
Text: "Donald Trump is a hoe"
Label: "polarized": 1

Example 5:
Text: "Fake news, begone English propaganda!"
Label: "polarized": 1
   
Text: {text}
"""
\end{verbatim}
}

Prompt 12: This prompt further extends the example-based prompt by including reasoning explanations for each example. The model is instructed to reason before answering, making this prompt the most explicit in terms of both definition and reasoning guidance.
{\small
\begin{verbatim}
PROMPT = """Given the following texts in X 
language, classify them as polarized or 
non-polarized based on the following 
definition: 

Definition of Polarization: Polarization 
refers to the phenomenon where opinions, 
beliefs, or behaviors become more extreme 
or divided, resulting in increased separation 
and conflict between different groups. These 
groups may be based on politics or ideology, 
ethnicity or race, religion, gender, or sexual 
orientation. Polarization involves negative 
attitudes such as stereotyping, denigration, 
dehumanization, de-individualization, or 
intolerance toward individuals or groups outside 
one's own, while exhibiting blind support and 
solidarity toward those within the group. 

Reason with the text before answering and answer 
only in one number ("1" if polarized or "0" if 
non-polarized).

Examples -
Text: Black immigrants should leave the USA. 
Reasoning: This statement stereotypes a group 
based on race and immigration status, implying a 
negative attitude and 
exclusion. 
Label: 1

Text: Africans are lazy because they are backward. 
Reasoning: Negative trait assignment based on 
ethnicity, showing polarization. 
Label: 1

Text: Muslims are terrorists. 
Reasoning: Promotes harmful stereotypes and 
dehumanization. 
Label: 1

Text: African women are beautiful. 
Reasoning: Positive generalization, not polarizing. 
Label: 0

Text: {Text}
"""
\end{verbatim}
}

\section{Appendix - Official Results and Ranking}

Table~\ref{tab:subtask3}, \ref{tab:subtask1}, \ref{tab:subtask2} shows complete official results along with comparison with Polar Baseline. The highlighted F1 scores indicates the performance relative to the Polar Baseline. In most of the languages across all the subtasks our system shows much better performance compared to Baseline.

\begin{table}[htbp]
\centering
\caption{Performance of Lingo Research Group vs. POLAR Baseline in Subtask 3}
\label{tab:subtask3}
\begin{tabular}{lccc}
\toprule
\textbf{Language} & \textbf{Lingo F1} & \textbf{Baseline F1} & \textbf{Rank} \\
\midrule
amh & \textbf{0.5190} & 0.4433 & -- \\
arb & \textbf{0.5329} & 0.3902 & -- \\
ben & \textbf{0.1609} & 0.0868 & -- \\
deu & \textbf{0.4494} & 0.3485 & -- \\
eng & \textbf{0.4686} & 0.4100 & -- \\
fas & \textbf{0.3259} & 0.2004 & -- \\
hau & 0.1111 & \textbf{0.7456} & -- \\
hin & \textbf{0.7186} & 0.2348 & -- \\
khm & 0.3136 & \textbf{0.6095} & -- \\
nep & \textbf{0.6685} & 0.1314 & 3rd \\
ori & 0.2490 & \textbf{0.3841} & -- \\
pan & \textbf{0.4972} & 0.4561 & -- \\
spa & 0.4706 & \textbf{0.5088} & -- \\
swa & \textbf{0.4844} & 0.2205 & -- \\
tel & 0.3451 & \textbf{0.6738} & -- \\
tur & 0.4656 & \textbf{0.7693} & -- \\
urd & \textbf{0.5606} & 0.5316 & -- \\
zho & \textbf{0.6535} & 0.0000 & -- \\
\bottomrule
\end{tabular}
\end{table}

\begin{table*}[htbp]
\centering
\scriptsize
\setlength{\tabcolsep}{2.5pt}
\renewcommand{\arraystretch}{0.9}
\resizebox{\textwidth}{!}{%
\begin{tabular}{lcccc@{\hspace{0.6cm}}lcccc@{\hspace{0.6cm}}lcccc}
\cmidrule(r){1-5} \cmidrule(r){6-10} \cmidrule(r){11-15}
\multicolumn{5}{c}{\textbf{Subtask 1}} & \multicolumn{5}{c}{\textbf{Subtask 2}} & \multicolumn{5}{c}{\textbf{Subtask 3}} \\
\cmidrule(r){1-5} \cmidrule(r){6-10} \cmidrule(r){11-15}
\textbf{Team} & \textbf{Total} & \textbf{1st} & \textbf{2nd} & \textbf{3rd} &
\textbf{Team} & \textbf{Total} & \textbf{1st} & \textbf{2nd} & \textbf{3rd} &
\textbf{Team} & \textbf{Total} & \textbf{1st} & \textbf{2nd} & \textbf{3rd} \\
\cmidrule(r){1-5} \cmidrule(r){6-10} \cmidrule(r){11-15}
\cellcolor{blue3}UTokyo Tsuruoka Lab & 12 & 8 & 4 & 0 & \cellcolor{blue3}UTokyo Tsuruoka Lab & 13 & 7 & 5 & 1 & \cellcolor{blue3}SMASH & \multicolumn{1}{r}{16} & \multicolumn{1}{r}{9} & \multicolumn{1}{r}{4} & \multicolumn{1}{r}{3} \\
\cellcolor{blue2}NYCU-NLP & 12 & 3 & 5 & 4 & \cellcolor{blue2}NYCU-NLP & 15 & 6 & 5 & 4 & \cellcolor{blue2}NYCU-NLP & \multicolumn{1}{r}{11} & \multicolumn{1}{r}{7} & \multicolumn{1}{r}{3} & \multicolumn{1}{r}{1} \\
\cellcolor{blue1}PSK & 9 & 2 & 4 & 3 & \cellcolor{blue1}SMASH & 13 & 4 & 6 & 3 & \cellcolor{blue1}Sagarmatha & \multicolumn{1}{r}{4} & \multicolumn{1}{r}{2} & \multicolumn{1}{r}{0} & \multicolumn{1}{r}{2} \\
CYUT & 4 & 2 & 0 & 2 & \cellcolor{lingoHL}\textbf{Lingo Research Group} & 7 & 2 & 1 & 4 & Ping An & \multicolumn{1}{r}{4} & \multicolumn{1}{r}{0} & \multicolumn{1}{r}{4} & \multicolumn{1}{r}{0} \\
SMASH & 7 & 1 & 2 & 4 & PolaFusion & 4 & 1 & 0 & 3 & PolaFusion & \multicolumn{1}{r}{7} & \multicolumn{1}{r}{0} & \multicolumn{1}{r}{2} & \multicolumn{1}{r}{5} \\
\cellcolor{lingoHL}\textbf{Lingo Research Group} & 5 & 1 & 1 & 3 & Sagarmatha & 2 & 1 & 0 & 1 & OZemi & \multicolumn{1}{r}{3} & \multicolumn{1}{r}{0} & \multicolumn{1}{r}{2} & \multicolumn{1}{r}{1} \\
taien & 3 & 1 & 1 & 1 & AIvengers & 1 & 0 & 1 & 0 & AIvengers & \multicolumn{1}{r}{4} & \multicolumn{1}{r}{0} & \multicolumn{1}{r}{1} & \multicolumn{1}{r}{3} \\
OZemi & 2 & 1 & 0 & 1 & ShefFriday & 1 & 0 & 1 & 0 & CYUT & \multicolumn{1}{r}{1} & \multicolumn{1}{r}{0} & \multicolumn{1}{r}{1} & \multicolumn{1}{r}{0} \\
Sagarmatha & 1 & 1 & 0 & 0 & Stochastic Gradient Descenders & 1 & 0 & 1 & 0 & ShefFriday & \multicolumn{1}{r}{1} & \multicolumn{1}{r}{0} & \multicolumn{1}{r}{1} & \multicolumn{1}{r}{0} \\
mdok-style & 1 & 1 & 0 & 0 & MSqrd & 1 & 0 & 1 & 0 & YEZE & \multicolumn{1}{r}{2} & \multicolumn{1}{r}{0} & \multicolumn{1}{r}{0} & \multicolumn{1}{r}{2} \\
PhatThachDau & 1 & 1 & 0 & 0 & CYUT & 1 & 0 & 0 & 1 & \cellcolor{lingoHL}\textbf{Lingo Research Group} & \multicolumn{1}{r}{1} & \multicolumn{1}{r}{0} & \multicolumn{1}{r}{0} & \multicolumn{1}{r}{1} \\
MKJ & 2 & 0 & 2 & 0 & YEZE & 1 & 0 & 0 & 1 &  &  &  &  &  \\
StanceLab & 2 & 0 & 2 & 0 & mdok-style & 1 & 0 & 0 & 1 &  &  &  &  &  \\
CUET-823 & 1 & 0 & 1 & 0 & YNU-HPCC & 1 & 0 & 0 & 1 &  &  &  &  &  \\
PolDeck & 1 & 0 & 1 & 0 & PolarMind & 1 & 0 & 0 & 1 &  &  &  &  &  \\
Projet Fil Rouge 821 & 1 & 0 & 1 & 0 &  & \multicolumn{1}{l}{} & \multicolumn{1}{l}{} & \multicolumn{1}{l}{} & \multicolumn{1}{l}{} &  &  &  &  &  \\
UMUSP & 1 & 0 & 1 & 0 &  & \multicolumn{1}{l}{} & \multicolumn{1}{l}{} & \multicolumn{1}{l}{} & \multicolumn{1}{l}{} &  &  &  &  &  \\
PolaFusion & 1 & 0 & 0 & 1 &  & \multicolumn{1}{l}{} & \multicolumn{1}{l}{} & \multicolumn{1}{l}{} & \multicolumn{1}{l}{} &  &  &  &  &  \\
YEZE & 1 & 0 & 0 & 1 &  & \multicolumn{1}{l}{} & \multicolumn{1}{l}{} & \multicolumn{1}{l}{} & \multicolumn{1}{l}{} &  &  &  &  &  \\
MoMo & 1 & 0 & 0 & 1 &  & \multicolumn{1}{l}{} & \multicolumn{1}{l}{} & \multicolumn{1}{l}{} & \multicolumn{1}{l}{} &  &  &  &  &  \\
Semantic Vectors & 1 & 0 & 0 & 1 &  & \multicolumn{1}{l}{} & \multicolumn{1}{l}{} & \multicolumn{1}{l}{} & \multicolumn{1}{l}{} &  &  &  &  &  \\
Tralaleros & 1 & 0 & 0 & 1 &  & \multicolumn{1}{l}{} & \multicolumn{1}{l}{} & \multicolumn{1}{l}{} & \multicolumn{1}{l}{} &  &  &  &  & \\

\cmidrule(r){1-5} \cmidrule(r){6-10} \cmidrule(r){11-15}
\end{tabular}%
}
\caption{Official Ranking - SemEval}
\label{tab:medal_by_task}
\end{table*}

\begin{table}[htbp]
\centering
\caption{Performance of Lingo Research Group vs. POLAR Baseline in Subtask 1}
\label{tab:subtask1}
\begin{tabular}{lccc}
\toprule
\textbf{Language} & \textbf{Lingo F1} & \textbf{Baseline} & \textbf{Rank} \\
\midrule
amh & \textbf{0.7928} & 0.7151 & 3rd \\
arb & 0.7690 & \textbf{0.7957} & -- \\
ben & 0.7925 & \textbf{0.8528} & -- \\
deu & \textbf{0.7322} & 0.6714 & -- \\
eng & \textbf{0.8098} & 0.7802 & -- \\
fas & 0.7191 & \textbf{0.8424} & 3rd \\
hau & 0.4719 & \textbf{0.7753} & -- \\
hin & \textbf{0.8212} & 0.7379 & 3rd \\
ita & 0.3459 & \textbf{0.6773} & -- \\
khm & \textbf{0.6645} & 0.6592 & -- \\
mya & \textbf{0.8773} & 0.8210 & -- \\
nep & \textbf{0.9180} & 0.8798 & 2nd \\
ori & \textbf{0.7975} & 0.7765 & -- \\
pan & \textbf{0.7946} & 0.7898 & 3rd \\
pol & \textbf{0.8431} & 0.7241 & 1st \\
rus & \textbf{0.7723} & 0.7457 & -- \\
spa & \textbf{0.7883} & 0.7266 & -- \\
swa & 0.7425 & \textbf{0.7571} & -- \\
tel & \textbf{0.7712} & 0.6440 & -- \\
tur & \textbf{0.7977} & 0.6957 & -- \\
urd & \textbf{0.8156} & 0.7890 & 3rd \\
zho & \textbf{0.9211} & 0.8691 & -- \\
\bottomrule
\end{tabular}
\end{table}

\begin{table}[htbp]
\centering
\caption{Performance of Lingo Research Group vs. POLAR Baseline in Subtask 2}
\label{tab:subtask2}
\begin{tabular}{lccc}
\toprule
\textbf{Language} & \textbf{Lingo F1} & \textbf{Baseline} & \textbf{Rank} \\
\midrule
amh & \textbf{0.5458} & 0.3716 & -- \\
arb & \textbf{0.6517} & 0.4855 & -- \\
ben & \textbf{0.4216} & 0.2887 & 1st \\
deu & \textbf{0.5994} & 0.4078 & 3rd \\
eng & \textbf{0.5027} & 0.3333 & -- \\
fas & \textbf{0.5757} & 0.4626 & -- \\
hau & \textbf{0.3022} & 0.2038 & -- \\
hin & 0.7704 & \textbf{0.7911} & -- \\
ita & 0.3409 & \textbf{0.3759} & -- \\
khm & \textbf{0.6939} & 0.6268 & -- \\
mya & \textbf{0.6772} & 0.4772 & -- \\
nep & \textbf{0.8047} & 0.7219 & 2nd \\
ori & 0.5394 & \textbf{0.5600} & -- \\
pan & \textbf{0.4555} & 0.3650 & -- \\
pol & \textbf{0.6253} & 0.4491 & 3rd \\
rus & \textbf{0.6023} & 0.5904 & -- \\
spa & \textbf{0.6640} & 0.5935 & -- \\
swa & \textbf{0.4929} & 0.4417 & -- \\
tel & \textbf{0.4054} & 0.3145 & -- \\
tur & \textbf{0.6242} & 0.4708 & 3rd \\
urd & \textbf{0.7978} & 0.7127 & 1st \\
zho & \textbf{0.8250} & 0.6697 & 3rd \\
\bottomrule
\end{tabular}
\end{table}

\end{document}